\let\oldvec\vec
\documentclass[draft,runningheads,citeauthoryear,oribibl]{llncs}
\let\vec\oldvec

\usepackage[T1]{fontenc}
\usepackage{graphicx}
\usepackage{amssymb}
\usepackage{amsmath}
\usepackage{algorithm}
\usepackage[noend]{algpseudocode}
\usepackage[group-separator={,}]{siunitx}

\newtheorem{mytheorem}{Theorem}
\newtheorem{mylemma}{Lemma}
\newtheorem{mydefinition}{Definition}
\newtheorem{mycorollary}{Corollary}
\newtheorem{myexample}{Example}

\newcommand{\myqed}{\mbox{$\diamond$}}

\begin{document}
\title{Greedy Algorithms for \\ Fair Division of Mixed Manna}
\titlerunning{Greedy Algorithms for Fair Division of Mixed Manna}

\author{Martin Aleksandrov \and Toby Walsh}
\authorrunning{M. Aleksandrov, T. Walsh}

\institute{Technical University Berlin, Germany \\
\email{\{martin.aleksandrov,toby.walsh\}@tu-berlin.de}}

\maketitle

\begin{abstract}
We consider a multi-agent model for fair division of mixed manna (i.e.\ items for which agents can have positive, zero or negative utilities), in which agents have additive utilities for bundles of items. For this model, we give several general impossibility results and special possibility results for three common fairness concepts (i.e.\ EF1, EFX, EFX$^3$) and one popular efficiency concept (i.e.\ PO). We also study how these interact with common welfare objectives such as the Nash, disutility Nash and egalitarian welfares. For example, we show that maximizing the Nash welfare with mixed manna (or minimizing the disutility Nash welfare) does not ensure an EF1 allocation whereas with goods and the Nash welfare it does. We also prove that an EFX$^3$ allocation may not exist even with identical utilities. By comparison, with tertiary utilities, EFX and PO allocations, or EFX$^3$ and PO allocations always exist. Also, with identical utilities, EFX and PO allocations always exist. For these cases, we give polynomial-time algorithms, returning such allocations and approximating further the Nash, disutility Nash and egalitarian welfares in special cases.
\end{abstract}


\section{Introduction}\label{sec:intro}

Fair division is the problem of allocating resources to agents \cite{steinhaus1948}. We consider a version of the problem with \emph{mixed manna}, in which the resources are \emph{indivisible} items and can be classified into three types: mixed goods, pure bads and dummy bads. The \emph{mixed goods} are items valued positively by some agents while valued non-positively by the other agents. These are \emph{goods} whenever they are valued non-negatively by all agents, and \emph{pure goods} whenever they are valued positively by all agents. The \emph{pure bads} are items valued negatively by all agents. The \emph{dummy bads} generate zero utilities for some agents and negative utilities for the other agents. Some related theoretical settings already use our model (e.g.\ ``good allocations'', ``chore allocations'' and ``allocation of objective goods and chores'') \cite{aziz2019gc}. Some related practical applications of our model include allocations of donated food products (e.g.\ \cite{aleksandrov2015ijcai}, https://www.foodbank.org.au/), allocations of time-based services (e.g.\ \cite{marks2012}, https://timebanks.org/) and matchings of tasks (e.g.\ \cite{ramshaw2012}, https://www.airtasker.com/). 

In this paper, we advance the ``state-of-the-art'' for the model of mixed manna by investigating how common normative properties of allocations such as EF1 (i.e.\ the envy of an agent is eliminated by removing \emph{some} item), EFX (i.e.\ the envy of an agent is eliminated by removing \emph{any} item) and PO (i.e.\ if some agent gets strictly higher utility in some other allocation, then some other agent gets strictly lower utility) interact with welfare objectives in these allocations such as the Nash welfare (i.e.\ the product of agents' utilities), the disutility Nash welfare (i.e.\ the product of agents' disutilities) and the egalitarian welfare (i.e.\ the minimum agent's utility). We also propose a new and stronger version of EFX, namely EFX$^3$, that requires EFX over all items, EFX only over the mixed goods and EFX only over the pure bads. We thus study further how this stronger fairness concept relates to the considered welfare objectives.

Existing works for fair division of mixed manna already study some of these interactions. Some of their approaches are successful and some others are unsuccessful. For example, in the setting with only goods, Caragiannis et al.\ \cite{caragiannis2016} proved that a natural modification of the Nash welfare maximizing rule satisfies both EF1 and PO. This appealing result suffers unfortunately from the high computational complexity of the rule. Indeed, computing allocations that maximize this welfare is $\mbox{NP}$-hard even in a simple setting such as the one with pure goods and identical additive utilities \cite{nguyen2013,ramezani2010,roos2010}. Perhaps, the most relevant observation for us is that these nice and elegant results break in our setting, supposing that some of the mixed goods are valued with strictly negative utilities by some of the agents. We give shortly examples for this observation.

Other prior works have considered optimizing other welfares in the case of bads such as the disutility Nash welfare. We believe that it seems natural to minimize this welfare as then agents would tend to receive bundles of bads they disvalue least. Unfortunately, this approach does not give us any fairness guarantees. Say, we wish to minimize this welfare over the set of all agents. It is easy to notice that each allocation that gives all bads to some agent minimizes the disutility Nash welfare. Clearly, such allocations falsify EF1. This is true even if we insist that each agent receives at least one bad. For example, Aziz et al.\ \cite{aziz2018gcv1} observed that minimizing this welfare under this assumption violates EF1 even when agents have identical utilities for bads, in which case each complete allocation of the bads is PO. 

An alternative approach is to look at allocations that maximize the disutility Nash welfare. One drawback of this approach is that each such allocation tends to give to agents bundles of their most disvalued bads. Nevertheless, such allocations are evidently the only ones that satisfy both EF1 and PO in some problems with bads. To see this, consider a problem with $n$ agents and $k\cdot n$ identical bads. It is easy to see that each allocation where some agent receives at least $(k+1)$ bads is not EF1 because there is some other agent with strictly less than $k$ bads. On the other hand, each EF1 alloction gives exactly $k$ bads to each agent and it is further maximal for the disutility Nash welfare. As a result, an algorithm returns an EF1 allocation in such problems iff it maximizes the disutility Nash welfare.

Maximizing the disutility Nash welfare is surprisingly related to maximizing the egalitarian welfare. To see this, consider again a problem with $n$ agents and $k\cdot n$ identical bads. An allocation that maximizes the egalitarian welfare gives to each agent exactly $k$ bads. As we argued previously, such an allocation maximizes the disutility Nash welfare. It is trivial to see that the opposite direction holds as well in such problems. 

In fact, some existing works consider the approximate envy-freeness guarantees of the maximal egalitarian allocations. For example, Plaut and Roughgarden \cite{plaut2018} prove that the {\em leximin} solution is EFX and PO with identical additive utilities. This solution selects the allocation which maximizes firstly the minimum utility and then, if there are multiple allocations which achieve that minimum utility, it chooses among those the allocation which maximizes the second minimum utility, and so on. Clearly, the leximin solution maximizes the egalitarian welfare. However, it is well-known that finding the leximin solution can require exponential time for general utilities \cite{dobzinski2013}, in which case maximizing the egalitarian welfare may not guarantee even EF1 (see \cite{caragiannis2016} for examples).

\section{Our results}\label{sec:results}

We advance the ``state-of-the-art'' in fair division of mixed manna by presenting several non-trivial contributions, primarily a number of general impossibility results and three polynomial-time algorithms for a number of special possibility results.

\begin{itemize}
\item \emph{General utilities}: With mixed goods, we show that maximizing the Nash welfare does not give us any EF1 or PO guarantees (Example~\ref{exp:one}). With pure bads, maximizing or minimizing the disutility Nash welfare is not related to EF1 (Examples~\ref{exp:two} and~\ref{exp:three}). Also, we prove that an EFX$^3$ allocation may not exist even with identical utilities (Example~\ref{exp:four}).\\
\item \emph{Tertiary utilities}: We give two algorithms, one for EFX and PO (Theorem~\ref{thm:tnashmax}) and another for EFX$^3$ and PO (Theorem~\ref{thm:tnashmaxmin}) allocations. Hence, an EFX$^3$ allocation exists in this domain. The EFX algorithm maximizes the egalitarian welfare (Lemma~\ref{lem:tnashmax}) whilst the EFX$^3$ algorithm may not do that (Lemma~\ref{lem:tnashmaxmin}).\\
\item \emph{Identical utilities}: We give an algorithm for EFX and PO (Theorem~\ref{thm:iminmax}) allocations. We further report a number of tight approximation guarantees of this algorithm wrt the Nash welfare with the pure goods and the disutility Nash welfare with pure bads (Lemmas~\ref{lem:iappgoods} and~\ref{lem:iappbads}).
\end{itemize} 

We conclude our paper with two observations: (1) our notion of EFX$^3$ belongs to a wider class of fairness concepts, (2) the Nash welfare of the leximin solution is $1.061$-approximation of the egalitarian welfare of the MNW solution. We believe that these observations open up interesting future directions.

\section{Related work}\label{sec:work}

One line of recent research studies various approximations of envy-freeness (i.e.\ EF) (see e.g.\ \cite{amanatidis2018,barman2018,lipton2004}). For instance, the idea of relaxing envy-freeness up to one item was for first time formally captured by means of EF1 in \cite{budish2011}. Later on, Caragiannis et al.\ \cite{caragiannis2016} proposed EFX that is weaker than EF but stronger than EF1. It remains an \emph{open} question if EFX allocations exist in general. Furthermore, for the model with mixed manna, Aziz et al.\ \cite{aziz2019gc} submitted the existence and computability of EFX allocations as ``an interesting line of research''. In our paper, we make a step forward and give two algorithms for EFX allocations with identical and tertiary utilities: the {\sc Nash-Max-Tertiary} and {\sc Max-Min-Identical} algorithms. In addition, we define a new fairness concept EFX$^3$, being stronger than EFX. With identical utilities, we show that EFX$^3$ allocations may not exist in some problems unlike EFX allocations. With tertiary utilities, we give an algorithm that returns an EFX$^3$ allocation: the {\sc Nash-Max-Min-Tertiary} algorithm. With just goods, our algorithms degenerate to some existing algorithms from Barman et al.\ \cite{barman2018}.

Another line of recent research investigates various welfares for fair division problems. For example, we mentioned the work of Caragiannis et al.\ \cite{caragiannis2016} who proved that maximizing the Nash welfare satisfies EF1 and PO for the case of goods. We show that this result ceases to hold for the case of mixed goods. Further, the product of disutilities was considered mainly in the context of divisible items (e.g.\ \cite{bogomolnaia2016gab,bogomolnaia2016gob}). In our work, we focus on indivisible items. The leximin solution, maximizing the egalitarian welfare, was developed as a metric of fairness in and of itself (see e.g.\ \cite{rawls1971,sen1976,sen1977}), and has been used before in fair division, though typically for randomized allocations (e.g.\ \cite{bogomolnaia2004}). Again, this one was used only for divisions of goods. We report some approximation guarantess of our algorithms wrt these welfare objectives. For example, the {\sc Max-Min-Identical} algorithm with just pure goods (bads) returns an allocation that is an $1.061$-approximation of the maximum geometric mean of the (disutility) Nash welfare. Moreover, the {\sc Nash-Max-Tertiary} algorithm maximizes the egalitarian welfare over all items. 

Some other recent works give polynomial-time algorithms for fair and efficient allocations. For example, Aziz et al.\ \cite{aziz2019gc} gave two such algorithms (i.e.\ the generalized Lipton's algorithm and the double round-robin algorithm) that return EF1 allocations in problems with mixed manna. Unfortunately, the allocations returned by them may not be PO even in the case of goods (see e.g.\ \cite{caragiannis2016,plaut2018}). Further, Barman, Murthy and Vaish \cite{barman2018fe} recently presented a pseudo-polynomial-time algorithm for computing an allocation that is EF1 and PO for goods. They show that this algorithm provides a polynomial-time $1.45$-approximation for the Nash welfare maximization problem. In fact, this is the best-known approximation factor for this problem. We add to these results as our algorithms require polynomial-time and also return EFX and PO allocations. Some more related works are \cite{aziz2016csc,bogomolnaia2019,bogomolnaia2017,caragiannis2012}.

\section{Model}\label{sec:model}

We consider a set $[n]=\lbrace 1,\ldots, n\rbrace$ of $n\in\mathbb{N}$ agents and a set $[m]=\lbrace 1,\ldots,m\rbrace$ of $m\in\mathbb{N}$ indivisible items. We let each $a\in [n]$ have some cardinal utility $u_a(o)\in\mathbb{R}$ for each $o\in [m]$. If $u_a(o)>0$, then $o$ is a \emph{good} for $a$. If $u_a(o)=0$, then $o$ is a \emph{dummy} for $a$. If $u_a(o)<0$, then $o$ is a \emph{bad} for $a$. We consider (1) \emph{mixed goods}\footnote{We say that the mixed goods are simply \emph{goods} whenever no agent has a strictly negative utility for any of them, and the goods are simply \emph{pure goods} whenever each agent has a strictly positive utility for each of them.} (i.e.\ items that are goods for some agents and not goods for the other agents), (2) \emph{pure bads} (i.e.\ items that are bads for all agents) and (3) \emph{dummy bads} (i.e.\ items that are dummies for some agents and bads for the other agents). We write $[m]^+=\lbrace o\in [m]|\exists a\in [n]:u_a(o)>0\rbrace$, $[m]^-=\lbrace o\in [m]|\forall a\in [n]:u_a(o)<0\rbrace$ and $[m]^0=\lbrace o\in [m]|\forall a\in [n]:u_a(o)\leq 0,\exists b\in [n]:u_b(o)=0 \rbrace$. We let $m^+$, $m^-$ and $m^0$ denote the cardinalities of $[m]^+$, $[m]^-$, and $[m]^0$.

For sets $N\subseteq [n]$ and $O\subseteq [m]$, an \emph{(complete) allocation} $A=(A_1,\ldots,A_n)$ of all items from $O$ to agents from $N$ is such that $\cup_{a\in N} A_a=O$ and $\cup_{a\in [n]\setminus N} A_a=\emptyset$. Thus, we write $\mathcal{A}(N,O)$ for the set of all complete allocations of the items from $O$ to the agents from $N$. We are mainly interested in complete allocations of all items from $[m]$. For each $a\in [n]$ and $M\subseteq [m]$, we write $u_a(M)$ for the \emph{bundle utility} of $a$ for the items in $M$. We write $u_a(o)$ for $u_a(\lbrace o\rbrace)$. We consider additive bundle utilities. That is, $u_a(M)=\sum_{o\in M} u_a(o)$. We say that these utilities are \emph{identical} iff, for each $M\subseteq [m]$, $u_a(M)=u_b(M)$ for each $a,b\in [n]$ and, for this reason, we write $u(M)$ in this case. We say that the cardinal utilities are \emph{tertiary} iff, for each $a\in [n]$ and each $o\in [m]$, $u_a(o)\in\lbrace -\alpha,0,\alpha \rbrace$ where $\alpha\in\mathbb{N}_{>0}$. 

\section{Axiomatic Properties}\label{sec:prel}

We next define formally some concepts that we use throughout the paper: envy-freeness, Pareto optimality and welfare optimality. 

\subsection{Envy-freeness}\label{subsec:fair}

We consider two approximations of envy-freeness of allocations \cite{foley1967}. These are EF1 and EFX. For goods, EF1 is proposed in \cite{budish2011} whereas EFX is from \cite{caragiannis2016}. However, for our model, we use their generalizations from \cite{aziz2019gc}.

\begin{mydefinition} $(${\em EF}$)$
An allocation $A$ is \emph{envy-free (EF)} if, for all $a,b\in [n]$, $u_a(A_a)\geq u_a(A_b)$.
\end{mydefinition} 

\begin{mydefinition} $(${\em EF1}$)$
An allocation $A$ is \emph{envy-free up to some item (EF1)} if, for all $a,b\in [n]$, either $u_a(A_a)\geq u_a(A_b)$ or $\exists o\in A_a\cup A_b$ such that $u_a(A_a\setminus\lbrace o\rbrace)\geq u_a(A_b\setminus\lbrace o\rbrace)$.
\end{mydefinition} 

\begin{mydefinition}$(${\em EFX}$)$
An allocation $A$ is \emph{envy-free up to any item (EFX)} if, for all $a, b\in [n]$, both of the two following conditions hold:
\begin{center}
\begin{enumerate}
\centering
\item $\forall o\in A_a$ such that $u_a(o)<0$: $u_a(A_a\setminus\lbrace o\rbrace)\geq u_a(A_b)$,
\item $\forall o\in A_b$ such that $u_a(o)>0$: $u_a(A_a)\geq u_a(A_b\setminus\lbrace o\rbrace)$.
\end{enumerate}
\end{center}
\end{mydefinition} 

We further define a stronger form of EFX, enforcing EFX over the items from $[m]^+$, $[m]^-$ and $[m]$. For a given $A=(A_1,\ldots,A_n)$, we let $A^+=(A^+_1,\ldots,A^+_n)$ and $A^-=(A^-_1,\ldots,A^-_n)$ where $A^+_a=A_a\cap [m]^+$ and $A^-_a=A_a\cap [m]^-$ for each $a\in [n]$.

\begin{mydefinition}$(${\em EFX$^3$}$)$
An allocation $A$ is \emph{EFX$^3$} if, for all $a, b\in [n]$, $A$ is EFX for $[m]$, $A^+$ is EFX for $[m]^+$ and $A^-$ is EFX for $[m]^-$.
\end{mydefinition}

\subsection{Pareto optimality}\label{subsec:eff}

We consider Pareto optimality as a measure of economic efficiency. Pareto optimality is proposed by Vilfredo Pareto long time ago in his seminal work \cite{pareto1896}. We next define this concept in our model.

\begin{mydefinition}$(${\em PO}$)$
An allocation $A$ is \emph{Pareto optimal (PO)} if there is no allocation $B$ such that $\forall a\in [n]$: $u_a(B_a)\geq u_a(A_a)$ and $\exists b\in [n]$: $u_b(B_b)> u_b(A_b)$.
\end{mydefinition} 

\subsection{Welfare optimality}\label{subsec:wel}

One well-studied social objective is the Nash welfare. The \emph{Nash welfare} (or \emph{NW}) in $A\in \mathcal{A}(N,O)$ is given by:

\begin{equation*}
\mbox{NW}(A)=\displaystyle\prod_{a\in N} u_a(A_a).
\end{equation*}

Another interesting social objective is the product of disutilities. The \emph{disutility Nash welfare} (or \emph{dNW}) in $A\in \mathcal{A}(N,O)$ is given by:

\begin{equation*}
\mbox{dNW}(A)=\displaystyle\prod_{a\in N} (-u_a(A_a)).
\end{equation*}

We also consider the minimum agent's utility in an allocation. The \emph{egalitarian welfare} (or \emph{EW}) in $A\in \mathcal{A}(N,O)$ is given by:

\begin{equation*}
\mbox{EW}(A)=\displaystyle\min_{a\in N} u_a(A_a).
\end{equation*}

We pay a minor attention in our work to allocations that are optimal or sub-optimal for a given welfare $\mbox{W}$. For this reason, we define such allocations.

\begin{mydefinition}$(${\em MW}$)$
An allocation $A\in \mathcal{A}(N,O)$ is \emph{maximal for the welfare $\mbox{W}$ (MW)} if, for each $B\in\mathcal{A}(N,O)$, $\mbox{\em W}(A)\geq\mbox{\em W}(B)$.
\end{mydefinition}

\begin{mydefinition}$(${\em $\beta$-approximation}$)$
For a fixed $\beta \in [0,1]$, an allocation $A\in \mathcal{A}(N,$ $O)$ is \emph{$\beta$-approximation of $\mbox{W}$} if, for each $B\in\mathcal{A}(N,O)$ that is maximal for $\mbox{W}$, $\mbox{\em W}(A)\geq \beta \cdot\mbox{\em W}(B)$.
\end{mydefinition} 

\section{General additive utilities}\label{sec:gen}

We begin with general utilities. With goods, it is well-known that maximizing the Nash welfare, over a largest set $N$ of agents to which one can simultaneously provide \emph{strictly positive} utilities\footnote{A set $N$ corresponds to the set of agents' vertices in a maximum matching in the bipartite graph $([n],E)$ where $(a,o)\in E$ iff $a\in [n]$, $o\in [m]^+$ and $u_a(o)>0$. Such a matching can be computed in $O(\max\lbrace n^{2.5},m^{2.5}\rbrace)$ time \cite{hopcroft1973}.}, selects an allocation that is EF1 and PO \cite{caragiannis2016}. Surprisingly, with mixed goods, these results do not hold for any of the MNW allocations from $\mathcal{A}(N,[m]^+)$. We show this in Example~\ref{exp:one}.

\begin{myexample}\label{exp:one}
Let us consider the following fair division problem with \num{2} agents and \num{3} mixed goods. Based on the below utilities, the largest set of agents who share positive utilities is $[2]$.

\begin{center}
\resizebox{0.5\columnwidth}{!}{
\begin{tabular}{|c|c|c|c|} \hline
  agent & mix. good 1 & mix. good 2 & pure good 3 \\ \hline
   1 & $2$ & $-4$ & $1$ \\
  2 & $-4$ & $2$ & $1$ \\ \hline
\end{tabular}
}
\end{center}

The allocations $A=(\lbrace 2,3\rbrace,\lbrace 1\rbrace)$ and $B=(\lbrace 2\rbrace,\lbrace 1,3\rbrace)$ maximize the Nash welfare, achieving a value of $12$. Neither $A$ nor $B$ is Pareto optimal because the agents are better off whenever they swap mixed goods 1 and 2. Further, $A$ and $B$ are not EF1. Indeed, $u_1(A_1\setminus\lbrace 2\rbrace)=1<2=u_1(A_2)$, $u_1(A_1\setminus\lbrace 3\rbrace)=-4<2=u_1(A_2)$ and $u_1(A_1)=-3<0=u_1(A_2\setminus\lbrace 1\rbrace)$. Similarly, agent 2 is not EF1 of agent 1 in the allocation $B$.
\myqed
\end{myexample}

With pure bads, we might wish to minimize the disutility Nash welfare over a largest set $M$ of agents to which one can simultaneously provide \emph{strictly negative} utilities\footnote{A set $M$ can be computed in $O(n)$ time. In fact, $M$ is a fixed subset of $m^-$ agents from $[n]$ if $m^-<n$ and $M=[n]$ if $m^-\geq n$.}. Unfortunately, there are problems where \emph{none} of the optimal allocations is even EF1. We demonstrate this in Example~\ref{exp:two}.

\begin{myexample}\label{exp:two}
Consider a simple fair division problem with \num{2} agents and \num{4} identical pure bads. The largest set of agents who share negative utilities is $[2]$. We note that each allocation is PO. 

\begin{center}
\resizebox{0.6\columnwidth}{!}{
\begin{tabular}{|c|c|c|c|c|} \hline
  agent & pure bad 1 & pure bad 2 & pure bad 1 & pure bad 2 \\ \hline
   1 & $-1$ & $-1$ & $-1$ & $-1$ \\
  2 & $-1$ & $-1$ & $-1$ & $-1$ \\ \hline
\end{tabular}
}
\end{center}

Each algorithm that minimizes the disutility Nash welfare would give zero bads to one of the agents. This allocation is not EF1. Each algorithm that minimizes the disutility Nash welfare, subject to the constraint that each agent should receive at least one pure bad, would give one pure bad to one of the agents and three pure bads to the other agent. Again, it should be clear that this allocation violates EF1.
\myqed
\end{myexample}

By Example~\ref{exp:two}, we conclude that there is \emph{no} algorithm that minimizes the disutility Nash welfare and returns an EF1 allocation. For this reason, we next attempt to maximize this welfare over a largest set $M$. As we show later, this approach gives us nice fairness guarantees in two special cases. However, in the general case, there are problems where both PO and EF1 may be falsified by each MdNW allocation from $\mathcal{A}(M,[m]^-)$. We illustrate this in Example~\ref{exp:three}.

\begin{myexample}\label{exp:three}
Again, consider a simple fair division problem with \num{2} agents and \num{3} pure bads. Given the following utilities, the largest set of agents who share negative utilities is $[2]$.

\begin{center}
\resizebox{0.5\columnwidth}{!}{
\begin{tabular}{|c|c|c|c|} \hline
  agent & pure bad 1 & pure bad 2 & pure bad 3 \\ \hline
   1 & $-2$ & $-1$ & $-4$ \\
  2 & $-1$ & $-2$ & $-4$ \\ \hline
\end{tabular}
}
\end{center}

The allocations $A=(\lbrace 1,3\rbrace,\lbrace 2\rbrace)$ and $B=(\lbrace 1\rbrace,\lbrace 2,3\rbrace)$ are MdNW with welfare of $12$. These are not Pareto optimal because agents are better off if they swap pure bads 1 and 2. Also, $A$ and $B$ are not EF1. Indeed, $u_1(A_1\setminus\lbrace 1\rbrace)=-4<-1=u_1(A_2)$ and $u_1(A_1\setminus\lbrace 3\rbrace)=-2<-1=u_1(A_2)$. Similarly, agent 2 is not EF1 of agent 1 in $B$.
\myqed
\end{myexample}

We next consider EFX$^3$. This property coincides with EFX whenever the problem contains just goods or just bads. It remains an \emph{open} question if EFX allocations exist in each problem with general utilities. However, EFX allocations do exist with identical utilities in these settings \cite{aleksandrov2018aef}. By comparison, in our setting with mixed manna, EFX$^3$ is strictly stronger than EFX and it may not be achievable even in this domain. We show this in Example~\ref{exp:four}.

\begin{myexample}\label{exp:four}
Suppose that there are \num{2} agents requesting \num{3} pure goods and \num{3} pure bads. We note that one EFX allocation gives items 1, 2, 3, 4 to agent 1 and items 5, 6 to agent 2.

\begin{center}
\resizebox{0.35\columnwidth}{!}{
\begin{tabular}{|c|c|c|c|c|} \hline

items & agent 1 & agent 2 \\ \hline
pure good 1 & $7$ & $7$ \\
pure good 2 & $6$ & $6$ \\
pure good 3 & $5$ & $5$ \\
pure bad 4 & $-100$ & $-100$ \\
pure bad 5 & $-2$ & $-2$ \\
pure bad 6 & $-1$ & $-1$ \\ \hline
\end{tabular}
}
\end{center}

However, there is no EFX$^3$ allocation. There are two EFX allocations of the pure goods and two EFX allocations of the pure bads: $A=(\lbrace 1\rbrace,\lbrace 2,3\rbrace)$, $B=(\lbrace 2,3\rbrace,\lbrace 1\rbrace)$, $C=(\lbrace 4\rbrace,\lbrace 5,6\rbrace)$ and $D=(\lbrace 5,6\rbrace,\lbrace 4\rbrace)$. If we unite (``agent-wise'') $A$ and $C$, then $u(A_1\cup C_1\setminus\lbrace 4\rbrace)=7<8=u(A_2\cup C_2)$. If we unite $A$ and $D$, then $u(A_2\cup D_2)=-89<-3=u(A_1\cup C_1\setminus\lbrace 1\rbrace)$. Hence, the unions of $A$ and $C$, or $A$ and $D$ are not EFX$^3$. Similarly, uniting $B$ and $C$, or $B$ and $D$ violates EFX$^3$. 
\myqed
\end{myexample}

In response to Examples~\ref{exp:one},~\ref{exp:two},~\ref{exp:three} and~\ref{exp:four}, we report two cases when EFX and PO can be achieved through maximization of the considered welfares: tertiary and identical utilities.

\section{Tertiary additive utilities}\label{sec:tertiary}

We continue with tertiary utilities. In this domain, it is easy to see that an agent is envy-free up to any item whenever they are envy-free up to some item. That is, EF1 degenerates to EFX. 

We, thus, give two algorithms for EFX and PO allocations. Both algorithms degenerate to an existing algorithm (i.e.\ the {\sc Alg-Binary} algorithm from \cite{barman2018}) for goods valued with utilities from $\lbrace 0,\alpha \rbrace$. As a result, they run in $O(2\cdot m\cdot (n+1)\cdot \ln(n\cdot m))$ time. However, they work also in our setting with utilities from $\lbrace -\alpha,0,\alpha \rbrace$. 

One of these algorithms returns an EFX and PO allocation, maximizing the egalitarian welfare but it does not bound the envy between agents for pure bads. The other one returns an EFX$^3$ and PO allocation, bounding by $\alpha$ this envy but it does not maximize the egalitarian welfare. 

\subsection{Algorithm for EFX and PO}\label{subsec:tefx}

We can compute in polynomial time an allocation that is MNW over the mixed goods, satisfying further EFX and PO. We next present a simple algorithm for this task.

\begin{algorithm}
\caption{Nash-Max-Tertiary($[n],[m],(u_a(o))_{n\times m}$)}\label{alg:efxter}
\begin{algorithmic}[1]
\Procedure{Nash-Max-Tertiary}{$[n],[m],(u_a(o))_{n\times m}$}
\State $\forall a\in [n]: A_a\gets\emptyset$
\If{$[m]^+\neq\emptyset$} 
\State $N\gets$ a largest set given $[m]^+$
\State $I\gets([m]^+,\emptyset,\ldots,\emptyset)$ 
\State $A\gets${\sc Alg-Binary}($N,[m]^+,(u_a(o))_{n\times m},I$) 
\EndIf
\For{$o\in [m]^-$} 
\State $\mbox{MaxUtil}(A)\gets \lbrace  a\in [n]|\forall b\in [n]:u_a(A_a)\geq u_b(A_b)\rbrace$
\State pick $a\in\mbox{MaxUtil}(A)$
\State $A_a\gets A_a\cup\lbrace o\rbrace$
\EndFor
\For{$o\in [m]^0$}
\State pick $a\in\lbrace b\in [n]|u_b(o)=0\rbrace$
\State $A_a\gets A_a\cup\lbrace o\rbrace$
\EndFor
\State \Return $A$
\EndProcedure
\end{algorithmic}
\end{algorithm}

Algorithm~\ref{alg:efxter} works in three steps. Initially, in the first step, Algorithm~\ref{alg:efxter} computes a feasible subset $N$\footnote{Barman et al.\ \cite{barman2018} assumed that $N=[n]$. In our work, we drop this assumption. For this reason, we need to compute one such set at line \num{4} in Algorithm~\ref{alg:efxter}.} and an allocation $A$ over $N$ that is MNW for the items from $[m]^+$. For this purpose, Algorithm~\ref{alg:efxter} calls the existing {\sc Alg-Binary} algorithm from \cite{barman2018}. 

Afterwards, in the second step, Algorithm~\ref{alg:efxter} allocates greedily the items from $[m]^-$ to agents, diminishing the utility of each agent with maximum utility in $A$ until all of these utilities become equal to the second maximum utility of an agent in $A$, supposing that there are enough items in $[m]^-$. In this way, Algorithm~\ref{alg:efxter} computes another allocation and repeats this process with this allocation until all items from $[m]^-$ are allocated. 

At the end, in the third step, Algorithm~\ref{alg:efxter} allocates each item from $[m]^0$ to an agent who values it with zero utility, returning a complete allocation of all items from $[m]$. We next prove that this one is EFX and PO.

\begin{mytheorem}\label{thm:tnashmax}
With tertiary additive utilities, the {\sc Nash-Max-Tertiary} algorithm returns an EFX and PO allocation.
\end{mytheorem}

\begin{myproof} {\bf Items from $[m]^+$ (Lines 3-6 in Algorithm~\ref{alg:efxter})}: The allocation of items from $[m]^+$ is done by calling the {\sc Alg-Binary} algorithm\footnote{The {\sc Alg-Binary} algorithm constructs a directed graph in which there is a vertex for each agent $a\in N$ and there is an directed edge from $a\in N$ to $b\in N$ iff there is an item $o\in [m]^+$ with $u_b(o)=\alpha>0$ and $o$ is currently allocated to agent $a$. It follows that the algorithm's output would not change supposing the negative utilities of agents from $[n]$ for items from $[m]^+$ are replaced by zero utilities.}. This algorithm takes as input $N$, $[m]^+$ and some allocation $I$ that is complete for $[m]^+$ (e.g.\ see line \num{5} in Algorithm~\ref{alg:efxter}), and returns as output an MNW allocation $A$ over $N$ for $[m]^+$. Hence, $A$ is EFX and PO by the result of Caragiannis et al.\ \cite{caragiannis2016}.

{\bf Items from $[m]^-$ (Lines 7-10 in Algorithm~\ref{alg:efxter})}: The allocation of items from $[m]^-$ proceeds in rounds. In round 1, we let $\sigma(A)$ denote the utility order induced by $A$. Wlog, $u_1(A_1)\geq u_2(A_2)\geq\ldots\geq u_n(A_n)$. In this round, all agents with maximum utility in $A$ (i.e.\ $u_1(A_1)$) receive items from $[m]^-$ in a round-robin fashion until the utility of each of these agents reaches the second utility level in $\sigma(A)$. This process may terminate before this happens if there are not enough items in $[m]^-$ for this to happen. At the end of round 1, the algorithm returns some allocation $B_1$. If there are unallocated items in $[m]^-$, the algorithm proceeds to the next round.

In round $k>1$, we let $B^{k-1}$ denote the partial allocation returned after the first $(k-1)$ rounds and $\sigma(B^{k-1})$ the utility order induced by $B^{k-1}$. Wlog, $u_1(B^{k-1}_1)\geq u_2(B^{k-1}_2)\geq\ldots\geq u_n(B^{k-1}_n)$. In this round, all agents with maximum utility in $B^{k-1}$ (i.e.\ $u_1(B^{k-1}_1)$) receive items from $[m]^-$ in a round-robin fashion until the utility of each of these agents reaches the second utility level in $\sigma(B^{k-1})$, obtaining another allocation $B^k$ at the end of round $k$. For each round $k$, it is easy to see that the allocation $B^k$ is EFX and PO because $A$ is EFX and PO, $B^{k-1}$ is EFX and PO and, at round $k$, items from $[m]^-$ are allocated in a round-robin fashion to agents with maximum utility in $\sigma(B^{k-1})$. 

{\bf Items from $[m]^0$ (Lines 11-14 in Algorithm~\ref{alg:efxter})}:
From an EFX perspective, it is best if each item from $[m]^0$ is allocated to an agent who has zero utility for it. Moreover, the sum of agents' utilities in the final allocation $A$ is $(m^+-m^-)\cdot \alpha$. Hence, PO follows. Otherwise, this sum in a Pareto-improvement of $A$ would be strictly more than $(m^+-m^-)\cdot \alpha$.
\myqed
\end{myproof}

We further prove that Algorithm~\ref{alg:efxter} returns an allocation that maximizes the egalitarian welfare over all items.

\begin{mylemma}\label{lem:tnashmax}
With tertiary additive utilities, the algorithm {\sc Nash-Max-Tertiary} returns an MEW allocation.
\end{mylemma}

\begin{myproof}
Let us consider the MNW allocation $A$ returned by {\sc Alg-Binary}. We argue that $A$ maximizes the egalitarian welfare over $N$ for $[m]^+$. By PO, it follows $u_a(A_a)>0$ for each $a\in N$. Let us assume that there is some other allocation $B$ such that $\mbox{EW}(B)>\mbox{EW}(A)$. Furthermore, let us consider agent $a=\arg\min_{c\in N} u_c(A_c)$. By fixing $A_a$ and varying the items in the bundle $\cup_{b\in N\setminus\lbrace a\rbrace} A_b$ among agents in $N\setminus\lbrace a\rbrace$, we cannot obtain an allocation with a value of the egalitarian welfare that is greater than $\mbox{EW}(A)$. Therefore, as $\mbox{EW}(B)>\mbox{EW}(A)$, the allocation $B$ must be obtained by re-allocating the items in $A$, including those in $A_a$, in such a way so that $u_a(B_a)>u_a(A_a)$. It must be then the case that agent $a$ likes some items in $A_b$ for some $b\in N\setminus \lbrace a\rbrace$. We consider two cases for agent $b$. If $u_b(A_b)\leq u_a(A_a)+\alpha$, moving items from $A_b$ to $A_a$ results in an allocation with egalitarian welfare at most $\mbox{EW}(A)$. If $u_b(A_b)>u_a(A_a)+\alpha$,  agent $a$ does not like any item in $A_b$. Otherwise, moving one such item from $A_b$ to $A_a$ increases the Nash welfare under $A$, contradicting its maximality under $A$. Consequently, as $\mbox{EW}(B)>\mbox{EW}(A)$, the allocation $B$ must be obtained by removing items from $A_a$ and/or moving items to $A_a$ from agents with utility in $A$ that is at most $u_a(A_a)+\alpha$. Wlog, let us remove the subset of items $S$ from $A_a$. It follows that moving items to $A_a\setminus S$ from agents with utility in $A$ that is at most $u_a(A_a)+\alpha$ gives us an allocation with egalitarian welfare at most $\mbox{EW}(A)$. Therefore, it must be the case that $\mbox{EW}(B)\leq\mbox{EW}(A)$ holds. We reached a contradiction with our assumption. Finally, let us next consider the allocations of items from $[m]^-$ and $[m]^0$. From an egalitarian perspective, the algorithm allocates optimally each of these items. The result follows.
\myqed
\end{myproof}

\subsection{Algorithm for EFX$^3$ and PO}\label{subsec:tefxefx}

We observed examples of problems where the allocation returned by Algorithm~\ref{alg:efxter} gave all pure bads to a single agent simply because they received large enough utility for the mixed goods. We believe that such an allocation might be perceived as unfair as it does not bound the envy between agents for pure bads.

By comparison, an MdNW allocation of the pure bads bounds by $\alpha$ this envy. Additionally, such an allocation maximizes the egalitarian welfare over these items because it gives to each agent $\lfloor m^-/n\rfloor$ or $\lfloor m^-/n\rfloor+1$ pure bads and each agent has disutility $-\alpha$ for each pure bad.

In fact, we can compute in polynomial time an allocation of all items whose sub-allocation over the mixed goods is MNW and sub-allocation over the pure bads is MdNW, satisfying further EFX$^3$ and PO. Such an allocation bounds by $\alpha$ the envy between agents for all items from $[m]$, all mixed goods from $[m]^+$ and all pure bads from $[m]^-$, as well as maximizes the egalitarian welfares over $[m]^+$ and $[m]^-$. We present below an algorithm for this task.

\begin{algorithm}
\caption{Nash-Max-Min-Tertiary($[n],[m],(u_a(o))_{n\times m}$)}\label{alg:efxefxter}
\begin{algorithmic}[1]
\Procedure{Nash-Max-Min-Tertiary}{$[n],[m],(u_a(o))_{n\times m}$}
\State $\forall a\in [n]: A_a\gets\emptyset,B_a\gets\emptyset,C_a\gets\emptyset$
\If{$[m]^+\neq\emptyset$} 
\State $N\gets$ a largest set given $[m]^+$
\State $I\gets([m]^+,\emptyset,\ldots,\emptyset)$ 
\State $A\gets${\sc Alg-Binary}($N,[m]^+,(u_a(o))_{n\times m},I$) 
\EndIf
\If{$[m]^-\neq\emptyset$} 
\State $M\gets$ a largest and feasible set given $N$ and $[m]^-$
\State $\sigma_{M,A}\gets$ a max-min priority order given $M$ and $A$
\State $B\gets${\sc Round-Robin}($\sigma_{M,A},[m]^-,(u_a(o))_{n\times m},A$) 
\EndIf
\For{$o\in [m]^0$}
\State pick $a\in\lbrace b\in [n]|u_b(o)=0\rbrace$
\State $C_a\gets B_a\cup\lbrace o\rbrace$
\EndFor
\State \Return $C$
\EndProcedure
\end{algorithmic}
\end{algorithm}

Algorithm~\ref{alg:efxefxter} makes use of two largest subsets $N$ and $M$\footnote{See footnotes 2 and 3.}, subject to the following \emph{feasibility} constraint: $M\subseteq N$ if $m^-\leq |N|$ and, otherwise, $N\subseteq M$. This one is crucial for the EFX guarantees of the returned allocation. Indeed, if $N\cap M=\emptyset$, then this allocation may violate EFX. We illustrate this in Example~\ref{exp:five}.

\begin{myexample}\label{exp:five}
Let us consider the below problem and the largest subsets $N=\lbrace 1,2\rbrace$ and $M=\lbrace 3\rbrace$. We note that both sets $N$ and $M$ are well-defined.

\begin{center}
\resizebox{0.5\columnwidth}{!}{
\begin{tabular}{|c|c|c|c|} \hline
  agent & pure good 1 & pure good 2 & pure bad 3 \\ \hline
   1 & $1$ & $1$ & $-1$  \\
  2 & $1$ & $1$ & $-1$  \\ 
  3 & $1$ & $1$ & $-1$  \\ \hline
\end{tabular}
}
\end{center}

Algorithm~\ref{alg:efxefxter}, with $N$ and $M$ that do not satisfy the feasibility constraint, returns $A=(\lbrace 1\rbrace,\lbrace 2\rbrace,\lbrace 3\rbrace)$ or $B=(\lbrace 2\rbrace,\lbrace 1\rbrace,\lbrace 3\rbrace)$. Clearly, $A$ is not EF1 (and, therefore, EFX): $u_3(A_3)=-1<0=u_3(A_1\setminus\lbrace 1\rbrace)=u_3(A_2\setminus\lbrace 2\rbrace)$ and $u_3(A_3\setminus\lbrace 3\rbrace)=0<1=u_3(A_1)=u_3(A_2)$. Similarly, $B$ violates EFX as well.
\myqed
\end{myexample}

Algorithm~\ref{alg:efxefxter} initializes three different allocations $A$, $B$ and $C$ with empty bundles and further executes two ``if'' conditions and one ``for'' loop. In the first ``if'' condition, Algorithm~\ref{alg:efxefxter} computes an MNW allocation $A$ of the items from $[m]^+$, using the existing {\sc Alg-Binary} algorithm. 

In the second ``if'' condition, Algorithm~\ref{alg:efxefxter} extends $A$ to another allocation $B$ over some largest and feasible set $M$, that is MdNW and PO for the items from $[m]^-$. For this purpose, Algorithm~\ref{alg:efxefxter} allocates the items from $[m]^-$ in a round-robin fashion, according to a \emph{max-min priority order} $\sigma_{M,A}$\footnote{Wlog, let $M=\lbrace 1,\ldots,l \rbrace$ and $u_1(A_1)\geq\ldots\geq u_l(A_l)$. A max-min priority order is an order of the agents from $M$ that is consistent with the order induced by their utilities in $A$. For example, $(1,\ldots,l)$ is one such order in which agent 1 has the \emph{max}imum utility and agent $l$ has the \emph{min}imum utility in $A$.}. 

In the ``for'' loop, Algorithm~\ref{alg:efxefxter} allocates each item from $[m]^0$ to an agent who values it with zero utility, extending $B$ to an allocation $C$. Algorithm~\ref{alg:efxefxter} terminates by returning the allocation $C$. We next prove that this returned allocation is EFX$^3$ and PO.

\begin{mytheorem}\label{thm:tnashmaxmin}
With tertiary additive utilities, the {\sc Nash-Max-Min-Tertiary} algorithm returns an EFX$^3$ and PO allocation.
\end{mytheorem}

\begin{myproof} {\bf Items from $[m]^+$ (Lines 3-6 in Algorithm~\ref{alg:efxefxter})}: As we discussed in Theorem~\ref{thm:tnashmax}, the allocation $A$ returned by the {\sc Alg-Binary} algorithm is EFX and PO for the items from $[m]^+$.

{\bf Items from $[m]^-$ (Lines 7-10 in Algorithm~\ref{alg:efxefxter})}: Wlog, $[m]^-=\lbrace 1,\ldots,j\rbrace$. Let these items be allocated from $1$ to $j$. For $k$ from \num{0} to $j$, we let $A^{k}$ denote the partial allocation of the first $k$ of these $j$ items, extending $A$. We note that  $A^{0}=A$ and $A^{j}=B$ hold.

{\em Step 1. Pareto optimality of $A^k$}: We argue that the allocation $A^{k}$ is PO for each fixed $k\in [j]$. The sum of agents' utilities in each PO allocation of the items from $[m]^+$ and the first $k$ items from $[m]^-$ is $m^+\cdot\alpha-k\cdot\alpha$. This is also true for $A^{k}$. Hence, this allocation must be PO. Otherwise, the total utility in an Pareto improvement of $A^k$ would be strictly higher than $m^+\cdot\alpha-k\cdot\alpha$. 

{\em Step 2. EFX of $A^k$}: We next argue that $A^{k}$ is EFX by induction on $k$. In the base case, let $k$ be \num{0}. The result follows. In the hypothesis case, we assume that $A^{(k-1)}$ is EFX. In the step case, let us consider round $k$. Wlog, let $M=\lbrace 1,\ldots,l\rbrace$, $u_1(A_1)\geq\ldots\geq u_l(A_l)$ and $\sigma=\sigma_{M,A}=(1,\ldots,l)$. 

Wlog, let it be agent $a(k)$'s turn in $\sigma$ to pick item $k$ from $[m]^-$. By the choice of $\sigma$, we next prove that the following property holds.

\begin{quote}
\emph{Property ($\star$)}: In $A^{(k-1)}$, any agent $b$ before agent $a(k)$ in $\sigma$ is EFX, agent $a(k)$ is EF and any agent $a$ after agent $a(k)$ in $\sigma$ is EFX. 
\end{quote}

In $A^{(k-1)}$, for some $s\in\mathbb{N}_{>0}$, let agent $b$ hold $s$ of the first $(k-1)$ items whilst agents $a(k)$ and $a$ hold each $(s-1)$ of these items. 

For agents $b$ and $c$ before $b$ in $\sigma$, $u_b(A^{(k-1)}_b)=u_b(A_b)-s\alpha\geq u_b(A_c)-\alpha-s\alpha$ by the EFX of $A^{(k-1)}$. Agent $c$ gets $s$ bads in this allocation. Hence, $u_b(A_c)-\alpha-s\alpha=u_b(A^{(k-1)}_c)-\alpha$. For agents $b$ and $c$ after $b$ in $\sigma$, $u_b(A^{(k-1)}_b)=u_b(A_b)-s\alpha\geq u_c(A_c)-s\alpha$. Agent $c$ receives either $s$ or $(s-1)$ bads in $A^{(k-1)}$. Hence, $u_c(A_c)-s\alpha\geq u_c(A^{(k-1)}_c)-\alpha$. By the PO of $A^{(k-1)}$, it follows $u_c(A^{(k-1)}_c)\geq u_b(A^{(k-1)}_c)$. Agent $b$ is EFX in $A^{(k-1)}$. 

For agents $a$ and $c$ before $a$ in $\sigma$, $u_a(A^{(k-1)}_a)=u_a(A_a)-(s-1)\alpha\geq u_a(A_c)-\alpha-(s-1)\alpha$ by the EFX of $A^{(k-1)}$. Agent $c$ receives either $s$ or $(s-1)$ bads in this allocation. In either case, $u_a(A_c)-\alpha-(s-1)\alpha\geq u_a(A^{(k-1)}_c)-\alpha$. For agents $a$ and $c$ after $a$ in $\sigma$, $u_a(A^{(k-1)}_a)=u_a(A_a)-(s-1)\alpha\geq u_c(A_c)-(s-1)\alpha$. By the PO of $A$, it follows $u_c(A_c)\geq u_a(A_c)$. Agent $c$ gets $(s-1)$ bads in $A^{(k-1)}$. Consequently, $u_c(A_c)-(s-1)\alpha\geq u_a(A_c)-(s-1)\alpha=u_a(A^{(k-1)}_c)$. Agent $a$ is EFX in $A^{(k-1)}$.

For agents $a(k)$ and $b$, $u_{a(k)}(A^{(k-1)}_{a(k)})=u_{a(k)}(A_{a(k)})-(s-1)\alpha\geq u_{a(k)}(A_b)-\alpha-(s-1)\alpha$ by the EFX of $A^{(k-1)}$. Agent $b$ gets $s$ bads in this allocation. Consequently, it follows $u_{a(k)}(A_b)-\alpha-(s-1)\alpha=u_{a(k)}(A^{(k-1)}_b)$. For agents $a(k)$ and $a$, $u_{a(k)}(A^{(k-1)}_{a(k)})=u_{a(k)}(A_{a(k)})-(s-1)\alpha\geq u_a(A_a)-(s-1)\alpha$. By the PO of $A$, it follows $u_{a}(A_{a})\geq u_{a(k)}(A_a)$. Agent $a$ gets $(s-1)$ bads in $A^{(k-1)}$. Therefore, $u_a(A_a)-(s-1)\alpha\geq u_{a(k)}(A_a)-(s-1)\alpha=u_{a(k)}(A^{(k-1)}_a)$. Thus, unlike agents $a$ and $b$, agent $a(k)$ is EF in $A^{(k-1)}$.

Property ($\star$) holds. Let us next consider the agents inside $M$. As agent $b$ is EFX in $A^{(k-1)}$, it follows that they remain EFX in $A^{k}$. As agent $a(k)$ is envy-free in $A^{(k-1)}$, it follows that they become EFX in $A^{k}$ after they receive item $k$ as they value this item with $-\alpha$. As agent $a$ is EFX in $A^{(k-1)}$, it follows that they remain EFX in $A^{k}$. Also, consider the agents outside $M$. Each agent $b\in [n]\setminus M$ receives no bads. Hence, they are EFX in $A^{k}$ because they are EFX in $A$. Hence, $A^{k}$ is EFX. 

We also note that the allocation $(B_1\setminus A_1,\ldots,B_n\setminus A_n)$ is EFX over the items from $[m]^-$. Moreover, it is PO for these items as the sum of agents' utilities in it is $-m^-\cdot \alpha$. This is the maximal sum possible for these items.

{\bf Items from $[m]^0$ (Lines 11-14 in Algorithm~\ref{alg:efxefxter})}: As we discussed in Theorem~\ref{thm:tnashmax},  it is best if each item from $[m]^0$ is allocated to an agent who has zero utility for it. Hence, the final allocation $C$ is EFX and PO within the set $\mathcal{A}([n],[m])$ of complete allocations.
\myqed
\end{myproof}

One downside of Algorithm~\ref{alg:efxefxter} is that it may not maximize the egalitarian welfare over all items from $[m]$ (unlike Algorithm~\ref{alg:efxter} in Lemma~\ref{lem:tnashmax}). This follows because Algorithm~\ref{alg:efxefxter} may give to each agent the same number of pure bads. In fact, there are problems where each EFX$^3$ and PO allocation does that. As a result, each algorithm that returns such an allocation would fail to maximize this welfare as it might be more optimal from an egalitarian perspective to give all pure bads to the agent with the greatest utility over the mixed goods.

\begin{mylemma}\label{lem:tnashmaxmin}
With tertiary additive utilities, there is \emph{no} algorithm that returns an EFX$^3$, PO and MEW allocation.
\end{mylemma}

\begin{myproof}
The proof is via a simple counter-example. Let us consider a fair division problem with \num{2} agents, \num{2} goods and \num{2} pure bads.

\begin{center}
\resizebox{0.6\columnwidth}{!}{
\begin{tabular}{|c|c|c|c|c|} \hline
  agent & good 1 & good 2 & pure bad 3 & pure bad 4 \\ \hline
   1 & $1$ & $1$ & $-1$  & $-1$ \\
  2 & $0$ & $0$ & $-1$  & $-1$ \\  \hline
\end{tabular}
}
\end{center}

Each algorithm that returns an PO allocation would give both goods to agent 1. Each algorithm that returns an EFX$^3$ allocation would give one pure bad to agent 1 and the other pure bad to agent 2. We note that Algorithm~\ref{alg:efxefxter} does exactly this. The value of the egalitarian welfare in such an allocation is equal to $-1$. However, the maximal egalitarian allocation gives both goods to agent 1 and also both pure bads to agent 1, achieving a welfare of $0$.
\myqed
\end{myproof}

We note that Lemma~\ref{lem:tnashmaxmin} holds further for algorithms that are guaranteed to return an allocation which satisfies EFX$^3$ but may falsify PO. For completeness, we summarize this result.

\begin{mycorollary}\label{cor:tnashmax}
With tertiary additive utilities, there is \emph{no} algorithm that computes an EFX$^3$ and MEW allocation.
\end{mycorollary}

\section{Identical additive utilities}\label{sec:ident}

We end with identical utilities. In this domain, Pareto optimality follows trivially because the sum of the agents' utilities in each complete allocation is equal to $\sum_{o\in [m]} u(o)$. 

By Example~\ref{exp:four}, an EFX$^3$ allocation may not exist. For this reason, we give just one algorithm that returns an EFX allocation. This algorithm degenerates to an existing algorithm (i.e.\ the {\sc Alg-Identical} algorithm from \cite{barman2018}) for pure goods, that returns an EFX allocation. However, our algorithm returns such an allocation in fair division of mixed manna. 

Furthermore, it has been proven that maximizing exactly the Nash welfare in a problem with pure goods or the disutility Nash welfare in a problem with pure bads selects an EFX allocation \cite{aleksandrov2018aef}. Nevertheless, computing such an allocation is in $\mbox{NP}$-hard \cite{aziz2018gcv1,roos2010}. For this reason, we also consider the approximation guarantees of our algorithm wrt these welfares in these cases. 

\subsection{Algorithm for EFX and PO}\label{subsec:iefx}

Interestingly, we can compute in $O(m\cdot n)$ time an EFX allocation. We next give one algorithm for this task. Algorithm~\ref{alg:efxident} allocates the items one-by-one in a sequence of non-increasing absolute utility. If the current item is a bad, then Algorithm~\ref{alg:efxident} gives it to an agent who has currently \emph{max}imum utility. Otherwise, Algorithm~\ref{alg:efxident} gives it to an agent who has currently \emph{min}imum utility. 

\begin{algorithm}
\caption{Max-Min-Identical($[n],[m],(u(o))_n$)}\label{alg:efxident}
\begin{algorithmic}[1]
\Procedure{Max-Min-Identical}{$[n],[m],(u(o))_n$} 
\State $\forall a\in [n]: A_a\gets\emptyset$
\State $\sigma\gets (1,\ldots,m)$ \Comment{wlog, $|u(1)|\geq\ldots\geq |u(m)|$}
\For{$j\gets 1:m, o\gets \sigma(j)$} 
\If {$o$ is a bad} 
\State $\mbox{MaxUtil}(A)\gets \lbrace  a\in [n]|\forall b\in [n]:u(A_a)\geq u(A_b)\rbrace$
\State pick $a\in\mbox{MaxUtil}(A)$
\EndIf
\If {$o$ is a good or a dummy} 
\State $\mbox{MinUtil}(A)\gets \lbrace a\in [n]|\forall b\in [n]:u(A_a)\leq u(A_b)\rbrace$
\State pick $a\in\mbox{MinUtil}(A)$
\EndIf
\State $A_a\gets A_a\cup\lbrace o\rbrace$
\EndFor
\State \Return $A$ 
\EndProcedure
\end{algorithmic}
\end{algorithm}

\begin{mytheorem}\label{thm:iminmax}
With identical additive utilities, the {\sc Max-Min-Identical} algorithm returns an EFX and PO allocation.
\end{mytheorem}

\begin{myproof}
Let us consider the returned allocation. We prove EFX by induction on $o\in [m]$, supposing that $\sigma=(1,\ldots,m)$ is such that $|u(1)|\geq\ldots\geq |u(m)|$. In the base case, let $o$ be $1$. The allocation of item $o$ is clearly EFX. In the hypothesis, let $o$ be item $(j-1)$ and suppose that the partially constructed allocation $A=(A_1,\ldots,A_n)$ is EFX. In the step case, let $o$ be item $j$.

\emph{Case 1}: $o$ is a good or a dummy. WLOG, let agent $1\in\mbox{MinUtil}(A)$ and the algorithm gives $o$ to agent $1$. We note that $u(o)\geq 0$. Consider $B=(B_1,\ldots,B_n)$ with $B_1=A_1\cup\lbrace o\rbrace$ and $B_a=A_a$ for each $a\in [n]\setminus\lbrace 1\rbrace$. We prove that $B$ is EFX. For this purpose, we only need to show that, for each $a\in [n]\setminus\lbrace 1\rbrace$, agent $a$ is EFX of agent 1 and agent 1 is EFX of agent $a$. For each other pair of agents, this result follows by the induction hypothesis.

\emph{$1\rightarrow a$}: We have that $u(B_1)=u(o)+u(A_1)\geq u(A_1)$. By the hypothesis, $u(A_1)\geq u(A_a\setminus\lbrace g\rbrace)$ for each $g\in A_a$ with $u(g)>0$ and $u(A_1\setminus\lbrace b\rbrace)\geq u(A_a)$ for each $b\in A_1$ with $u(b)<0$. Therefore, $u(B_1)\geq u(A_1)\geq u(A_a\setminus\lbrace g\rbrace)=u(B_a\setminus\lbrace g\rbrace)$ and $u(B_1\setminus\lbrace b\rbrace)\geq u(A_a\setminus\lbrace b\rbrace)\geq u(A_a)=u(B_a)$ as $B_a=A_a$. 

\emph{$a\rightarrow 1$}: We have that $u(A_a)\geq u(A_1)$ as agent $1\in\mbox{MinUtil}(A)$. Hence, agent $a$ is envy-free of agent $1$ in $A$ and also $u(B_a)\geq u(B_1\setminus\lbrace o\rbrace)$ as $B_a=A_a$ and $B_1\setminus\lbrace o\rbrace=A_1$. As the items are revealed in a non-increasing absolute value, $u(g)\geq u(o)$ for each $g\in A_1$ with $u(g)>0$. Hence, $u(B_1)\geq u(B_1\setminus\lbrace o\rbrace)\geq u(B_1\setminus\lbrace g\rbrace)$ for each $g\in A_1$ with $u(g)>0$. We also have that $u(A_a\setminus\lbrace b\rbrace)> u(A_1)$ for each $b\in A_a$ with $u(b)<0$. Again, due to the order of the items, $|u(b)|\geq u(o)$. Therefore, $u(B_a\setminus\lbrace b\rbrace)\geq u(B_a)+u(o)=u(A_a)+u(o)$ as $B_a=A_a$. In the last step, we have $u(B_a\setminus\lbrace b\rbrace) \geq u(A_1)+u(o)=u(B_1)$.

\emph{Case 2}: $o$ is a bad. WLOG, let agent $1\in\mbox{MaxUtil}(A)$ and the algorithm gives $o$ to agent $1$. We note that $u(o)<0$. Consider $B=(B_1,\ldots,B_n)$ with $B_1=A_1\cup\lbrace o\rbrace$ and $B_a=A_a$ for each $a\in [n]\setminus\lbrace 1\rbrace$. We prove that $B$ is EFX. We show this only for agent $1$ and agent $a\in [n]\setminus\lbrace 1\rbrace$.

\emph{$1\rightarrow a$}: We have that $u(A_1)\geq u(A_a)$ as $1\in\mbox{MaxUtil}(A)$. Hence, agent $1$ is envy-free of agent $a$ in $A$ and also $u(B_1\setminus\lbrace o\rbrace)\geq u(B_a)$ as $B_a=A_a$ and $B_1\setminus\lbrace o\rbrace=A_1$. As the items are revealed in a non-increasing absolute value, $|u(b)|\geq |u(o)|$ for each $b\in A_1$ with $u(b)<0$. Hence, $u(B_1\setminus\lbrace b\rbrace)\geq u(B_1\setminus\lbrace o\rbrace)\geq u(B_a)$ for each $b\in A_1$ with $u(b)<0$. We also have that $u(A_1)> u(A_a\setminus\lbrace g\rbrace)$ for each $g\in A_a$ with $u(g)>0$. Again, due to the order of the items, $u(g)\geq |u(o)|$. Therefore, $u(B_a\setminus\lbrace g\rbrace)\leq u(B_a)-|u(o)|=u(A_a)-|u(o)|$ as $B_a=A_a$. And, we have $u(A_a)-|u(o)|\leq u(A_1)-|u(o)|=u(B_1)$. Consequently, $u(B_1)\geq u(B_a\setminus\lbrace g\rbrace)$ for each $g\in B_a$ with $u(g)>0$. 

\emph{$a\rightarrow 1$}: We have that $u(B_1)=u(o)+u(A_1)\leq u(A_1)$. By the hypothesis, $u(A_a)\geq u(A_1\setminus\lbrace g\rbrace)$ for each $g\in A_1$ with $u(g)>0$ and $u(A_a\setminus\lbrace b\rbrace)\geq u(A_1)$ for each $b\in A_a$ with $u(b)<0$. Therefore, $u(B_a)=u(A_a)\geq u(A_1\setminus\lbrace g\rbrace)\geq u(B_1\setminus\lbrace g\rbrace)$ as $u(A_1)\geq u(B_1)$ and $u(B_a\setminus\lbrace b\rbrace)= u(A_a\setminus\lbrace b\rbrace)\geq u(A_1)\geq u(B_1)$ because $B_a=A_a$. 

The algorithm returns a complete allocation. As we discussed previously, each such allocation satisfies Pareto optimality.
\myqed
\end{myproof}

We proceed with the approximation guarantees of Algorithm~\ref{alg:efxident}. With pure goods, Barman et al.\ \cite{barman2018} proved that an arbitrary EFX allocation $A$ is an \num{1.061}-approximation of the geometric mean of the maximal Nash welfare. That is, $\sqrt[n]{\mbox{NW}(A)}\geq\frac{1}{1.061}\sqrt[n]{\mbox{NW}(A^{\mbox{\tiny MNW}})}$. By Theorem~\ref{thm:iminmax}, Algorithm~\ref{alg:efxident} returns such an allocation. We merely summarize this result for completeness.

\begin{mylemma}\label{lem:iappgoods}
With pure goods and identical additive utilities, the algorithm {\sc Max-Min-Identical} returns an \num{1.061}-approximation of the geometric mean of the maximal Nash welfare.
\end{mylemma}

Interestingly, with pure bads, Algorithm~\ref{alg:efxident} approximates further up to the same tight bound the geometric mean of the maximal disutility Nash welfare, i.e.\ $\sqrt[n]{\mbox{dNW}(A)}\geq\frac{1}{1.061}\sqrt[n]{\mbox{dNW}(A^{\mbox{\tiny MdNW}})}$.

\begin{mylemma}\label{lem:iappbads}
With pure bads and identical additive utilities, the algorithm {\sc Max-Min-Identical} returns an \num{1.061}-approximation of the geometric mean of the maximal disutility Nash welfare.
\end{mylemma}

\begin{myproof}
Consider a problem with bads. If $m^-<n$, the returned allocation is optimal. For this reason, let $m^-\geq n$. We write $\mathcal{P}$ for $([n],[m]^-,(u(o))_{o\in [m]^-})$ and $-\mathcal{P}$ for $([n],[m]^-,(v(o))_{o\in [m]^-})$ where $v(o)=-u(o)$ for each $o\in [m]^-$. We note that $-\mathcal{P}$ contains only pure goods. For each agent $a\in [n]$ and each bundle $S\subseteq [m]^-$, we let $v_a(S)$ denote the utility of $a$ for $S$. It follows that $v_a(S)=-u_a(S)$ due to additivity. 

Let us consider an allocation $B$ in $\mathcal{P}$. The disutility Nash welfare in $B$ is $\mbox{dNW}(B)=\prod_{a\in [n]} (-u_a(B_a))$. Further, we let $-B$ denote the same allocation but in $-\mathcal{P}$. That is, $-B_a=B_a$ and $v(-B_a)=-u(B_a)$ for each $a\in [n]$. We have $\mbox{NW}(-B)=\prod_{a\in [n]}v_a(-B_a)=\mbox{dNW}(B)$. It follows that the order induced by the disutility Nash welfare among the allocations in $\mathcal{P}$ coincides with the order induced by the Nash welfare among the allocations in $-\mathcal{P}$. As a result, an allocation is MdNW in $\mathcal{P}$ iff it is MNW in $-\mathcal{P}$.

We next prove that an allocation in $\mathcal{P}$ is EFX iff it is EFX in $-\mathcal{P}$. Let $-B$ be an EFX allocation in $-\mathcal{P}$ but suppose that $B$ is not EFX in $\mathcal{P}$. Hence, there are two agents, say 1 and 2, and one pure bad $o\in B_2$ such that $u(B_1\setminus\lbrace o\rbrace)< u(B_2)$. Therefore, $v(-B_2)<v(-B_1\setminus\lbrace o\rbrace)$ holds. It follows that agent 2 is not EFX of agent 1 for the pure good $o$ in $-B$. Hence, $-B$ is not EFX in $-\mathcal{P}$. This is a contradiction. Similarly, let $B$ be an EFX allocation in $\mathcal{P}$ but suppose that $-B$ is not EFX in $-\mathcal{P}$. We derive $v(-B_1)< v(-B_2\setminus\lbrace o\rbrace)$ and, therefore, $u(B_2\setminus\lbrace o\rbrace)<u(B_1)$. This contradicts the EFX of $B$ in $\mathcal{P}$.

Let $B$ be an arbitrary EFX allocation in $\mathcal{P}$ and $-B$ be its corresponding EFX allocation in $-\mathcal{P}$. We have $\mbox{dNW}(B)=\mbox{NW}(-B)$. By the result of Barman et al.\ \cite{barman2018}, $\sqrt[n]{\mbox{NW}(-B)}\geq\frac{1}{1.061}\sqrt[n]{\mbox{NW}(-B^{\mbox{\tiny OPT}})}$ where $-B^{\mbox{\tiny OPT}}$ is an MNW allocation in $-\mathcal{P}$. Let $B^{\mbox{\tiny OPT}}$ be its corresponding allocation in $\mathcal{P}$. Hence, $B^{\mbox{\tiny OPT}}$ is an MdNW allocation in $\mathcal{P}$ and $\mbox{NW}(-B^{\mbox{\tiny OPT}})=\mbox{dNW}(B^{\mbox{\tiny OPT}})$. We derive that $\sqrt[n]{\mbox{dNW}(B)}\geq\frac{1}{1.061}\sqrt[n]{\mbox{dNW}(B^{\mbox{\tiny OPT}})}$ holds for $B$ in $\mathcal{P}$. 

The result follows because the algorithm returns an EFX allocation by Theorem~\ref{thm:iminmax}.
\myqed
\end{myproof}

Barman et al.\ \cite{barman2018} gave an example\footnote{See Example 4.3 in \cite{barman2018}.} of a problem with pure goods, confirming that the approximation guarantee of an EFX ``good allocation'' wrt the geometric mean of the maximal Nash welfare is almost tight. We observe a similar finding for an EFX ``bad allocation'' and the geometric mean of the maximal disutility Nash welfare. The approximation factor of Lemma~\ref{lem:iappbads} is almost tight. We demonstrate this in Example~\ref{exp:six}.

\begin{myexample}\label{exp:six}
Consider a problem with \num{2} agents and $(m+2)$ pure bads, where $m$ is an even number. Further, we let the utilities of agents 1 and 2 for the items be identical: $-m,-m,-1,-1,\ldots,-1$. 

There is one allocation, say $A$, that gives both pure bads valued with $-m$ to one of the agents, say agent 1, and all other $m$ pure bads valued with $-1$ to agent 2. It is easy to see that this allocation satisfies EFX. Moreover, $\mbox{dNW}(A)=2m\cdot m=2m^2$.

Another EFX allocation, say $B$, gives to each agent one pure bad valued with $-m$ and $\frac{m}{2}$ pure bads valued with $-1$. This one maximizes the disutility Nash welfare and each agent receives the same disutility over the bads. Furthermore, $\mbox{dNW}(B)=\frac{3}{2}m\cdot \frac{3}{2}m=\frac{9}{4}m^2$. The approximation ratio of $A$ is:

\begin{equation*}
\frac{\sqrt{\mbox{dNW}(A)}}{\sqrt{\mbox{dNW}(B)}}=\sqrt{\frac{8}{9}}\approx\frac{1}{1.0607}.\myqed
\end{equation*}
\end{myexample}

\section{Future work}\label{sec:disc}

Our results enabled us to open up several interesting future directions. In this section, we discuss two of these prominent lines of research.

\subsection{New fairness concepts}\label{subsec:nfair}

The notion of EFX$^3$ is a member of a more general class of fairness concepts, imposing constraints on the sub-problems of mixed goods and pure bads in our setting. For example, we may insist on satisfying different fairness concepts in each of these sub-problems. More formally, this can be defined as follows.

\begin{mydefinition}$(${\em X-Y-Z-fairness}$)$
An allocation $A$ is \emph{X-Y-Z fair} if, for all $a, b\in [n]$, $A$ is $\mbox{X}$ for $[m]$, $A^+$ is $\mbox{Y}$ for $[m]^+$ and $A^-$ is $\mbox{Z}$ for $[m]^-$, where $\mbox{X}$, $\mbox{Y}$ and $\mbox{Z}$ are fairness properties.
\end{mydefinition}

Examples of fairness properties are EF1, EFX, PROP (proportionality), PROP1 (proportionality up to some item, see e.g.\ \cite{conitzer2017}), PROPX (proportionality up to any item, see e.g.\ \cite{moulin2019}), etc. From this perspective, EFX$^3$ is equivalent to EFX-EFX-EFX fairness. An interesting future direction is to design (polynomial-time) algorithms that satisfy some of these criteria in the general case.  

\subsection{New approximation results}\label{subsec:napp}

We can also relate some of our results to other existing approximation results. For example, as we mentioned earlier, Plaut and Roughgarden \cite{plaut2018} proved that the \emph{leximin} solution is EFX and PO with identical utilities and pure goods. This one returns an MEW allocation. By the result of Barman~\cite{barman2018}, it follows that $\sqrt[n]{\mbox{NW}(A^{\mbox{\tiny leximin}})}\geq\frac{1}{1.061}\sqrt[n]{\mbox{NW}(A^{\mbox{\tiny MNW}})}$ holds. Therefore, an leximin allocation approximates the maximal Nash welfare in this setting. We further note that $\sqrt[n]{\mbox{NW}(A^{\mbox{\tiny MNW}})}\geq \mbox{EW}(A^{\mbox{\tiny MNW}})$ and, therefore, $\sqrt[n]{\mbox{NW}(A^{\mbox{\tiny leximin}})}$ $\geq\frac{1}{1.061}\mbox{EW}(A^{\mbox{\tiny MNW}})$ hold, which gives us an appealing relation between the Nash welfare of an leximin allocation and the egalitarian welfare of an MNW allocation. In addition, $\mbox{NW}(A^{\mbox{\tiny EFX}})\geq \mbox{NW}(A^{\mbox{\tiny leximin}})$ holds for some EFX allocations because the set of leximin allocations is a subset of the set of EFX allocations. Hence, if $\sqrt[n]{\mbox{NW}(A^{\mbox{\tiny leximin}})}\geq\beta \sqrt[n]{\mbox{NW}(A^{\mbox{\tiny MNW}})}$ for some $\beta>\frac{1}{1.061}$, then we can derive a strictly better approximation factor than $1.061$. We submit the deeper study of these approximation guarantees as a promising direction.

\section{Conclusion}\label{sec:future}

We studied the problem of fairly allocating items in a multi-agent setting, supposing the items can be characterized into three categories: mixed goods, pure bads and dummy bads. We thus gave several general impossibility results in regard to common fairness concepts such as EF1, EFX and EFX$^3$. For example, maximizing the Nash welfare with mixed goods does not give us any EF1 guarantees. Also, minimizing the disutility Nash welfare is not related to EF1. And, an EFX$^3$ allocation may not exist even with identical utilities. 

Nevertheless, we identified several special cases when the considered fairness concepts can be achieved in combination with PO. For example, with tertiary utilities, we gave a polynomial-time algorithm for EFX and PO allocations and an algorithm for EFX$^3$ and PO allocations. As a second example, with identical utilities, we gave a linear-time algorithm for EFX and PO allocations. We also proved several approximation guarantees of our algorithms wrt to the Nash, disutility Nash and egalitarian welfares. Finally, we discussed our future directions.

\newpage

\bibliographystyle{splncs04}
\bibliography{arxiv}

\end{document}